\documentclass[letterpaper, 10 pt, conference]{ieeeconf}  

\IEEEoverridecommandlockouts                              

\usepackage{etoolbox} 
\usepackage[utf8]{inputenc} 
\usepackage[T1]{fontenc}    
\usepackage{fonttable}
\usepackage[english]{babel} 
\usepackage[protrusion=true,expansion=true]{microtype}
\usepackage{comment}
\usepackage{cancel}
\usepackage{csquotes}
\usepackage{listings}
\usepackage{nameref}
\usepackage{xspace}
\usepackage{setspace}
\usepackage{makeidx}
\usepackage{pdfpages}

\usepackage{times}
\let\labelindent\relax
\usepackage{enumitem}

\usepackage{amsmath,amssymb} 
\usepackage{amsfonts}       
\usepackage{mathtools}
\usepackage{array} 
\usepackage{nicefrac}       
\usepackage{breqn}          
\usepackage{textcomp}
\usepackage{bm} 
\usepackage{pifont}

\usepackage{graphicx}
\usepackage{wrapfig}
\usepackage{adjustbox} 
\usepackage{epsfig}

\usepackage{float}
\usepackage{subcaption}
\usepackage[font=small,labelfont=bf]{caption}
\usepackage{lscape}      

\usepackage{tikz}  
\usepackage{makecell}
\usepackage{overpic}
\usepackage{array} 
\usepackage{tabularx}
\usepackage{multirow}
\usepackage{multicol}
\usepackage{booktabs}   
\usepackage{tablefootnote}
\usepackage{tabularray}

\setlist[itemize]{noitemsep,leftmargin=*,topsep=0in}
\setlist[enumerate]{noitemsep,leftmargin=*,topsep=0in}

\usepackage[table]{xcolor}

\usepackage{xurl}

\usepackage{hyperref}
\hypersetup{
  pagebackref=false,
  breaklinks=true,
  colorlinks=true,
  urlcolor=blue,
  citecolor=blue,
  linkcolor=blue,
  bookmarks=false
}

\usepackage{blindtext}

\usepackage{lipsum}

\renewcommand{\paragraph}[1]{\vspace{0.2em}\noindent\textbf{#1} --}

\newcommand{\modelname}{\textsc{KeyGen}\xspace}

\title{\modelname: Unsupervised Keypoint based Object-Centric Representations \\ for Category-Level Policy Generalization}

\author{%
  Shuxin Cao$^{1}$\thanks{$^{1}$Georgia Institute of Technology \quad $^{2}$University of Toronto}, %
  Liquan Wang$^{1}$, %
  Masoud Moghani$^{2}$, %
  Benjamin Joffe$^{1}$, %
  Animesh Garg$^{1}$%
}

\begin{document}
\maketitle
\thispagestyle{empty}
\pagestyle{empty}

\newcommand{\shuxin}[1]{\textcolor{color1}{\mycomment{ansh: #1}}}
\newcommand{\ag}[1]{\textcolor{color2}{\mycomment{animesh: #1}}}
\newcommand{\licho}[1]{\textcolor{color3}{\mycomment{liquan: #1}}}
\newcommand{\masoud}[1]{\textcolor{color4}{\mycomment{masoud: #1}}}
\newcommand{\ben}[1]{\textcolor{color5}{\mycomment{Ben: #1}}}

%








\maketitle

\vspace{-11mm}



\begin{abstract}
Generalization in robotic manipulation requires policies to perform tasks across diverse unseen object instances that vary in shape, size, and pose. 
However, conventional behavior cloning (BC) methods often overfit to instance-specific geometry and appearance, limiting transfer to novel objects. 
We introduce \textbf{\modelname}, a framework that learns canonicalized semantic 3D keypoints from point clouds and uses them as structured object-centric representations for policy learning. 
A visuomotor diffusion policy conditions on these keypoints together with object-centric geometry to predict full manipulation trajectories, enabling consistent geometric correspondence across object instances. 
To evaluate category-level generalization, we construct a photorealistic simulation benchmark with three manipulation tasks and a planning-driven data generation pipeline that produces expert trajectories across diverse object instances. 
Experiments show that \modelname{} significantly outperforms prior methods on both seen and unseen objects under pose variation, scales effectively with additional demonstrations per object, maintains robustness to object rescaling, and achieves strong performance in both simulation and real-world manipulation.
Project page: \href{https://robo-keygen.github.io}{robo-keygen.github.io}.

\end{abstract}

\section{Introduction}
\label{sec:introduction}

Robot learning has achieved impressive progress through large-scale behavior cloning (BC)~\cite{brohan2023rt1roboticstransformerrealworld, brohan2023rt2visionlanguageactionmodelstransfer, shridhar2022perceiveractormultitasktransformerrobotic, goyal2023rvtroboticviewtransformer, ha2023scalingup, baku}. 
However, real-world manipulation requires operating over diverse object instances within the same category—such as cups, knives, or containers—that vary in shape, size, material, and pose. 
While BC performs well on training objects, it often fails to transfer to novel instances due to distribution shifts in geometry and appearance. 
Although recent large-scale robotic datasets~\cite{vuong2023open, walke2023bridgedata, khazatsky2024droid} have expanded data collection efforts, they remain orders of magnitude smaller than datasets in vision or NLP, and exhaustively covering deployment variability through data alone is infeasible. 
Enabling reliable generalization across intra-category variation therefore remains a fundamental challenge in robotic manipulation.

\begin{figure}[!t]
  \centering
  \includegraphics[page=11,width=0.99\linewidth]{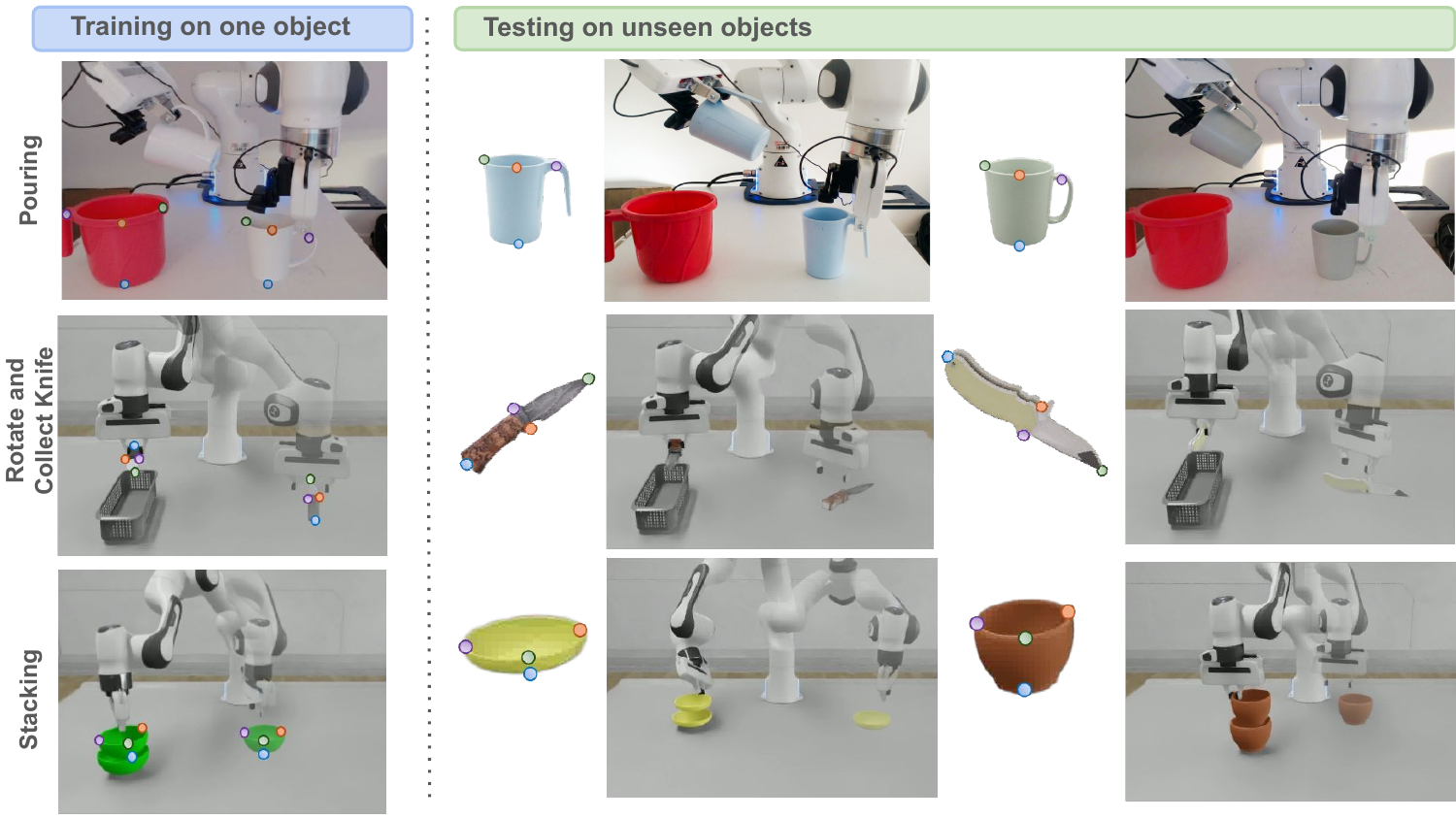}
  \caption{We propose \textbf{\modelname}, a framework that enables generalization across unseen object instances by leveraging semantically aligned 3D keypoints. \textbf{\modelname} uses unsupervised, task-agnostic keypoints as structured object representations, allowing policies to reason about geometry and pose for precise and transferable manipulation.}
  \label{fig:teaser}
\end{figure}

A common strategy for category-level generalization is to learn an intermediate object representation that suppresses instance-specific appearance while exposing manipulation-relevant geometry. 
Part-based priors from 2D segmentations, often guided by vision-language models~\cite{kerr2023, shen2023}, depend on accurate masks and frequently fail to produce consistent 3D correspondences under intra-category variation or clutter. 
Dense 3D features from pretrained visual encoders~\cite{wang2024, wang2024gendp} are largely appearance-driven and can miss the functional structure needed for contact-rich control, while keypoint-based one-shot imitation methods~\cite{duan2017oneshotimitationlearning} often rely on test-time planning or trajectory warping rather than learning transferable trajectory distributions. 
Motivated by these limitations, we propose \modelname{}, which grounds policy learning in semantically consistent 3D geometry for generalization to novel object instances.

To address this, \modelname{} self-supervisedly learns semantic 3D keypoints from multi-view RGB-D by canonicalizing object orientation and extracting a small set of object-centric landmarks from point clouds, yielding consistent geometric anchors that capture \textit{functional structure} and are robust to variations in shape, pose, and appearance. 
We train a diffusion-based behavior cloning policy conditioned on the object-centric point cloud and these keypoints to generate full continuous manipulation trajectories without external planning. 
Beyond representation, we introduce a photorealistic benchmark, a simulation environment plus data generation pipeline for category-level manipulation: it supports diverse object assets (e.g., mugs, knives, bowls) with systematic variation in geometry, pose, and scale, and uses task-specific finite state machines parameterized by canonical-frame geometric anchors to automatically produce expert trajectories across instances without per-object tuning.
This benchmark enables controlled evaluation of instance transfer, data efficiency, and robustness to geometric variation.

Building on this framework, we evaluate \modelname{} on photorealistic manipulation tasks with substantial intra-category variation and stress-test generalization along three practical axes: (i) transfer to unseen instances, (ii) data efficiency as demonstrations per object increase, and (iii) zero-shot robustness to object rescaling.
We further validate \modelname{} on real-world robot experiments, including pushing (push-shoe) and pouring, showing that the keypoint-conditioned policy remains effective under real-world sensing and dynamics. 
We summarize our contributions as follows:
\begin{itemize}
    \item \textbf{Representation.} A keypoint-based object-centric representation that canonicalizes orientation and extracts semantically consistent 3D keypoints from point clouds, robust to variations in shape, pose, and appearance.
    \item \textbf{Policy.} A DiT-based behavior cloning policy conditioned on keypoints and an object-centric 3D scene representation, generating dense action trajectories.
    \item \textbf{Simulation \& Data.} A photorealistic simulation suite and a planning-driven data generation pipeline that automatically produces expert trajectories across diverse object instances, enabling controlled evaluation of instance transfer, data efficiency, and scale robustness.
\end{itemize}

\section{Related Work}
\label{sec:related-work}

\paragraph{Representations for Few-Shot Generalization in Manipulation}
\label{subsec:generalizable-manipulation}
Recent work on generalizable manipulation often builds semantic 3D representations from pretrained vision models. D3Fields~\cite{wang2024} uses multi-view RGB-D with DINOv2 and SAM to construct 3D descriptor fields, and GenDP~\cite{wang2024gendp} conditions diffusion policies on DINO features for category-level transfer; related field-based methods such as LERF-TOGO~\cite{kerr2023} and DFFs~\cite{shen2023} embed CLIP/DINO features into NeRF-style representations for language-guided grasping. While effective, these approaches can require dense sensing or costly field construction and inherit the limitations of static, appearance-driven features. Another line of work incorporates geometric structure via equivariance: NDFs~\cite{simeonov2021} learn SE(3)-equivariant descriptors but rely on inference-time optimization, while EquivAct~\cite{yang2024equivact} and EquiBot~\cite{yang2024equibot} extend equivariance (e.g., to SIM(3)) and combine it with diffusion policies, which can be brittle in cluttered or multi-object scenes where global symmetry assumptions break. 
In contrast, we learn compact semantic 3D keypoints directly from multi-view RGB-D and use them as object-centric control tokens for policy learning.

\paragraph{Unsupervised 3D Keypoint Detection}
Semantic 3D keypoints support pose estimation, shape understanding, tracking, and manipulation, but supervised keypoint annotation on point clouds is expensive and often ambiguous, motivating unsupervised alternatives. Many methods assume objects are already in a canonical pose: Skeleton Merger~\cite{shi2021} learns keypoints and derives skeleton structure while enforcing proximity to the surface, USEEK~\cite{xue2023} adopts a teacher--student scheme with SE(3)-equivariant keypoints but can be challenged by symmetries, and Key-Grid~\cite{hou2024} uses autoencoding and reconstruction objectives to encourage geometric consistency. Other approaches address rotation variation without explicit equivariant networks by training with random rotations and enforcing alignment in canonical space, such as USIP~\cite{li2019} and SC3K~\cite{zohaib2023}; Canonical Capsules~\cite{sun2021} instead predicts a canonical pose as an intermediate representation to obtain consistent semantic keypoints. In contrast, our method targets raw multi-view RGB-D observations: we use a pretrained canonical orientation model to normalize object pose and then learn semantically consistent 3D keypoints in this canonical frame, enabling stable correspondences for downstream policy learning.

\section{\modelname: Approach}

\begin{figure*}
  \centering
  \includegraphics[page=12,width=\linewidth]{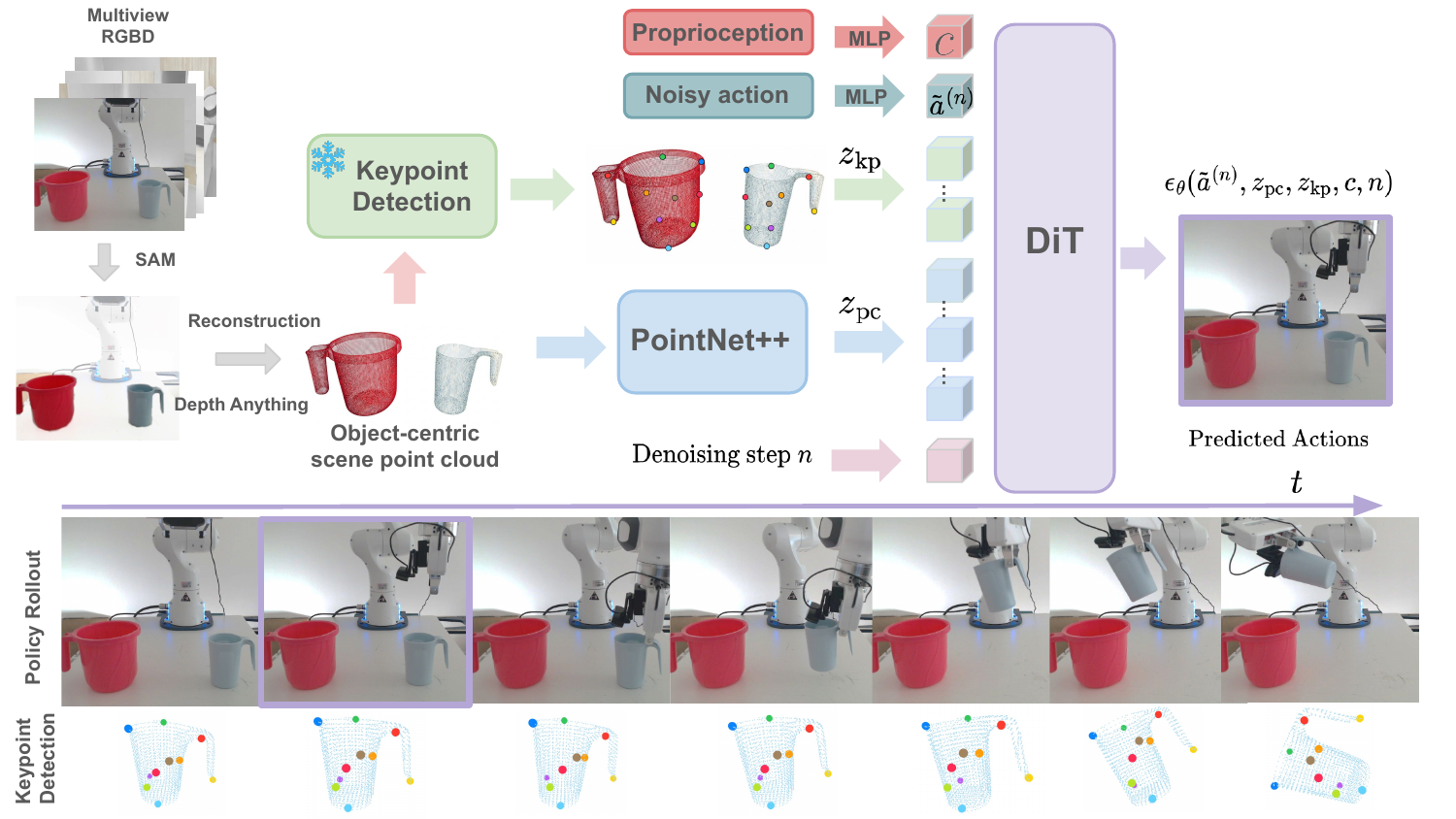}
  \caption{\textbf{Overview of KeyGen}. KeyGen segments multi-view RGBD images to create an object-centric point cloud, which is used to extract consistent semantic 3D keypoints. Together with proprioception and noisy action, the encoded object-centric point cloud and keypoints serve as conditioning features in the Diffusion Process. At inference time, the action is iteratively denoised to predict robot trajectory.}
  \label{fig:overview}
\end{figure*}

We formulate category-level manipulation as an MDP with state \(s \in \mathcal{S}\) (observations including object configuration and robot proprioception) and action \(a \in \mathcal{A}\) (robot control commands). 
Given demonstrations \(D=\{\tau_i\}_{i=0}^N\), where each trajectory \(\tau_i=(s_0,a_0,s_1,\ldots,a_T)\), we learn a behavior cloning policy \(\pi(a\mid s)\) that transfers from a limited set of training objects to novel instances within the same category.
Our objective is to match the expert action distribution by minimizing the discrepancy between \(P_{\text{pred}}(a\mid s)\) and \(P_{\text{gt}}(a\mid s)\), while leveraging category-level structure to handle variations in shape, size, and appearance. 
To support transfer, we use semantic 3D keypoints as the primary object-centric representation: a pretrained keypoint extractor produces instance-consistent landmarks, and the policy (Figure~\ref{fig:overview}) conditions on keypoints together with object-centric geometry to predict continuous action trajectories, capturing the temporal dynamics for precise manipulation beyond keyframe- or grasp-point-based formulations.

\subsection{3D Keypoint Detection}
\label{sec:keypoint_detect}
We present a self-supervised framework for 3D keypoint detection that generates task-agnostic, semantically meaningful, and geometrically consistent keypoints across object instances in the same category. To overcome pose ambiguity, where keypoints vary under different object orientations, we adopt a two-stage approach: first, we align each object to a canonical pose; then, we predict keypoints in this normalized frame. This design ensures the keypoints are both interpretable and pose-invariant, enabling better generalization for downstream policy learning.

\paragraph{Canonical orientation prediction}
Given an object-centric point cloud \(P \in \mathbb{R}^{3 \times N}\), we predict its canonical pose using a category-specific discrete rotation classifier over the 24-element octahedral group \(O = \{R_1, \dots, R_{24}\}\), inspired by~\cite{scarvelis2024}: $R^* = \arg\max_{R \in O} \Pr(R \mid P)$.

The classifier is trained on ShapeNet~\cite{chang2015shapenetinformationrich3dmodel}, where ground-truth canonical orientations are available. Discretizing rotation avoids instability from continuous regression and yields robust alignment. The input is then normalized as \(P_{\text{can}} = (R^*)^{-1} P\). The classifier is trained per category and kept fixed during keypoint learning.

\begin{figure*}[t]
    \centering
    \begin{minipage}{0.66\textwidth}
        \centering
        \includegraphics[width=\textwidth]{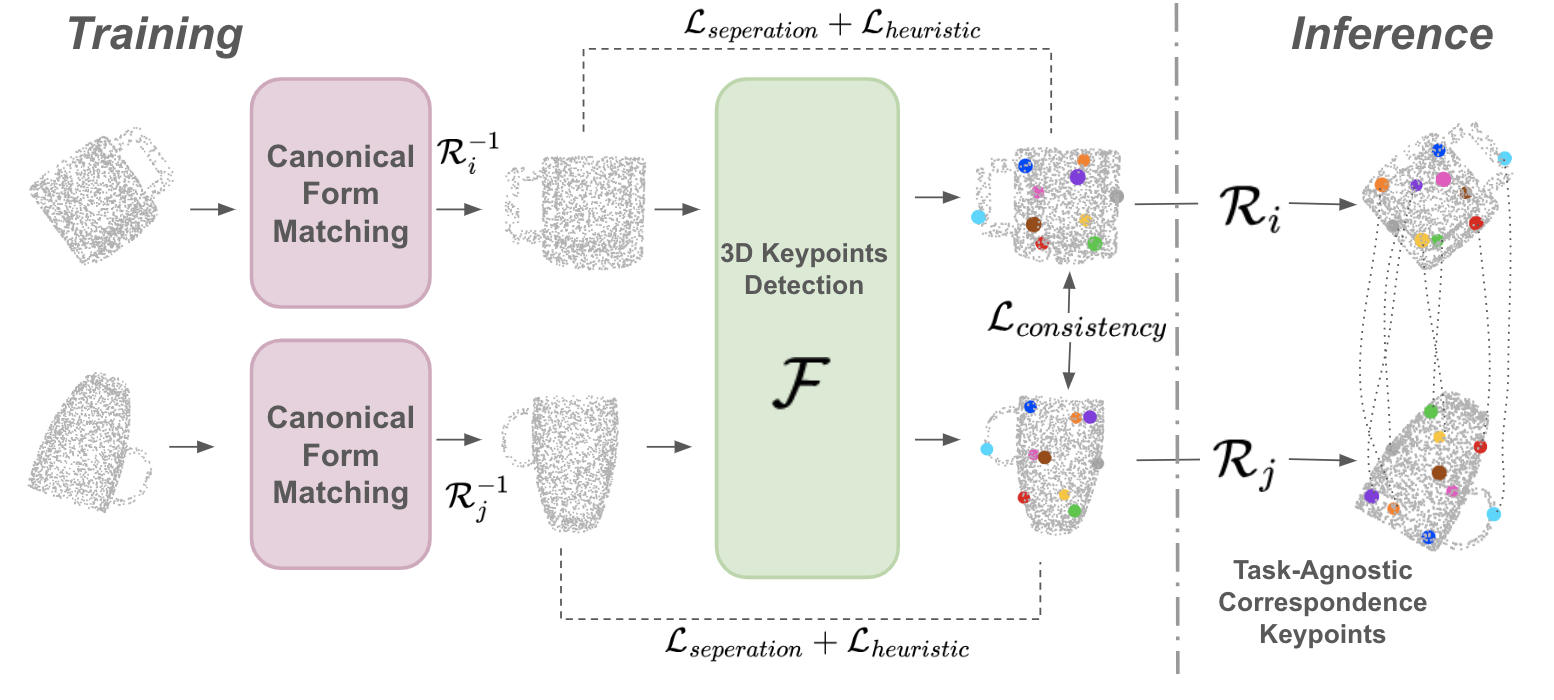}    
    \end{minipage}
    \begin{minipage}{0.32\textwidth}
        \centering
        \caption{\small \textbf{Category‐specific canonical alignment and 3D keypoint detection.} During training, two random partial scans of the same object are aligned to a shared canonical frame by a pretrained, category‐specific Canonical Form Matching network and then processed by the keypoint detector \(\mathcal{F}\); separation (\(\mathcal{L}_{\mathrm{sep}}\)), heuristic (\(\mathcal{L}_{\mathrm{heur}}\)), and consistency (\(\mathcal{L}_{\mathrm{cons}}\)) losses enforce landmark spread, saliency, and cross‐view repeatability. At inference, a single scan is aligned to canonical space, \(\mathcal{F}\) predicts keypoints, and they are reprojected to the original view via the estimated pose.}
        \label{fig:keypoint}
    \end{minipage}
\end{figure*}

\paragraph{Keypoint generation model training}
To learn view-consistent keypoints, we align two randomly sampled scans
\(P_a, P_b \in \mathbb{R}^{3\times N}\) to a shared canonical frame:
\begin{align}
P_a^{\text{can}} &= (R_a^*)^{-1} P_a, \qquad
P_b^{\text{can}} = (R_b^*)^{-1} P_b.
\end{align}
Each aligned cloud is passed through the keypoint detector \(\mathcal{F}\) to produce
\(K\) keypoints per view, parameterized as convex combinations of input points:
\begin{align}
\mathbf{k}_i &= \sum_{j=1}^N w_{ij}\,\mathbf{p}_j,
\quad w_{ij}\ge 0,\quad \sum_{j=1}^N w_{ij}=1.
\end{align}
Training minimizes a weighted sum of three terms,
\begin{align}
\mathcal{L}_{\mathrm{total}}
&= \lambda_{\mathrm{sep}}\,\mathcal{L}_{\mathrm{sep}}
 + \lambda_{\mathrm{heur}}\,\mathcal{L}_{\mathrm{heur}}
 + \lambda_{\mathrm{cons}}\,\mathcal{L}_{\mathrm{cons}},
\end{align}
where (i) \emph{separation} encourages spatial diversity, (ii) a \emph{heuristic} term biases
keypoints toward salient surface regions and coverage, and (iii) \emph{consistency} enforces
cross-view semantic alignment.

\noindent\textit{Separation.}
Let \(\mathcal{K}^{(a)}=\{\mathbf{k}^{(a)}_i\}_{i=1}^K\) and
\(\mathcal{K}^{(b)}=\{\mathbf{k}^{(b)}_i\}_{i=1}^K\).
With margin \(m_{\mathrm{sep}}>0\),
\begin{align}
\ell_{\mathrm{sep}}(\mathcal{K})
&= \binom{K}{2}^{-1}\!
   \sum_{i<j}
   \big[\,m_{\mathrm{sep}} - \lVert \mathbf{k}_i-\mathbf{k}_j\rVert_2\,\big]_+^{\,2},\\
\mathcal{L}_{\mathrm{sep}}
&= \tfrac{1}{2}\big(\ell_{\mathrm{sep}}(\mathcal{K}^{(a)})+\ell_{\mathrm{sep}}(\mathcal{K}^{(b)})\big),
\end{align}
\noindent\emph{where } \([x]_+=\max(0,x)\).

\noindent\textit{Heuristic (surface adherence + coverage).}
Let \(d(\mathbf{x},P)=\min_{\mathbf{p}\in P}\|\mathbf{x}-\mathbf{p}\|_2\) and
\(\mathrm{Vol}(\cdot)\) be the Axis-Aligned Bounding Box volume. With robust penalty \(\rho(\cdot)\)
(e.g., SmoothL1) and weights \(\alpha_{\mathrm{surf}},\alpha_{\mathrm{cov}}>0\),
\begin{align}
\ell_{\mathrm{surf}}(\mathcal{K},P)
&= \frac{1}{K}\sum_{i=1}^K d(\mathbf{k}_i,P),\\
\ell_{\mathrm{cov}}(\mathcal{K},P)
&= \rho\!\big(\mathrm{Vol}(\mathcal{K})-\mathrm{Vol}(P)\big),
\end{align}
\begin{align}
\mathcal{L}_{\mathrm{heur}}
&= \tfrac{1}{2}\alpha_{\mathrm{surf}}
 \Big[\ell_{\mathrm{surf}}(\mathcal{K}^{(a)},P_a^{\text{can}})
    + \ell_{\mathrm{surf}}(\mathcal{K}^{(b)},P_b^{\text{can}})\Big]\nonumber\\
&\quad + \tfrac{1}{2}\alpha_{\mathrm{cov}}
 \Big[\ell_{\mathrm{cov}}(\mathcal{K}^{(a)},P_a^{\text{can}})
    + \ell_{\mathrm{cov}}(\mathcal{K}^{(b)},P_b^{\text{can}})\Big].
\end{align}

\noindent\textit{Cross-view consistency.}
Reproject to the observation frames and penalize index-wise mismatch:
\begin{align}
\mathcal{L}_{\mathrm{cons}}
&= \frac{1}{K}\sum_{i=1}^K
 \big\|\,R_a^*\,\mathbf{k}_i^{(a)} - R_b^*\,\mathbf{k}_i^{(b)}\big\|_2^2.
\end{align}
This objective yields well-spread, surface-adherent, and view-invariant keypoints suitable
for downstream manipulation.

\paragraph{Keypoint inference}
At test time, a single object-centric point cloud \(P \in \mathbb{R}^{3 \times N}\) is first aligned to a canonical frame using the pretrained Canonical Form Matching network, yielding \(P_{\text{can}} = (R^*)^{-1} P\). The keypoint detector \(\mathcal{F}\) then predicts canonical keypoints \(\mathcal{K} = \mathcal{F}(P_{\text{can}})\), which are reprojected back to the original frame as \(\mathcal{K}_{\text{obs}} = R^* \mathcal{K}\). This two-stage process produces pose-invariant, semantically consistent 3D landmarks that abstract away viewpoint and geometric variations—providing a robust representation for generalizable manipulation policies.

\subsection{Policy Learning}
At each timestep $t$, we extract observations by rendering RGB-D images from four calibrated cameras uniformly placed around the workspace. 
We apply GroundingDINO~\cite{liu2024groundingdinomarryingdino} and SAM~\cite{sam} to segment task-relevant objects based on the task prompt, including both manipulated and target objects. 
The segmentation masks are used to filter the depth images, which are then fused across views to construct an object-centric point cloud $P_t \in \mathbb{R}^{N \times 3}$. 
A set of 3D keypoints $\mathcal{K}_t \in \mathbb{R}^{K \times 3}$ is inferred from $P_t$ using our pretrained keypoint detector. 
The robot's proprioceptive state $S_t \in \mathbb{R}^s$ (e.g., gripper pose) is also recorded. These three components—point cloud, keypoints, and robot state—form the observation set $C_t = \{P_t,\; \mathcal{K}_t,\; S_t\}$, which serves as the input to our diffusion-based policy model.

\paragraph{Condition Encoding.} Each modality is encoded independently using dedicated encoders. The point cloud is embedded via a PointNet-based encoder \cite{qi2017}: \( z_{\text{pc}} = f_{\text{pc}}(P_t) \). The keypoints are processed by a separate encoder: \( z_{\text{kp}} = f_{\text{kp}}(\mathcal{K}_t) \). The robot's proprioceptive state is encoded into a proprioception token \( c = f_{\text{grip}}(S_t) \). These representations serve as conditioning inputs to the diffusion model.

\paragraph{Diffusion Process.} We model the conditional distribution over action trajectories \( a_{1:T} \in \mathbb{R}^{T \times d} \) using a Denoising Diffusion Probabilistic Model (DDPM) \cite{ho2020denoising}. During training, we sample a clean trajectory \( a^{(0)} = a_{1:T} \) and a diffusion timestep \( n \sim \mathcal{U}(1, N) \), and perturb the trajectory with Gaussian noise \( \epsilon_n \sim \mathcal{N}(0, I) \) to obtain the noisy sample:$ \tilde{a}^{(n)} = \sqrt{\bar{\alpha}_n} a^{(0)} + \sqrt{1 - \bar{\alpha}_n} \epsilon_n $.

A noise prediction network \( \epsilon_\theta \) takes in \( \tilde{a}^{(n)} \), conditioning features, and timestep \( n \) to predict the added noise:
$ \hat{\epsilon}_n = \epsilon_\theta(\tilde{a}^{(n)}, z_{\text{pc}}, z_{\text{kp}}, c, n)$, 
and is trained using an $\ell_2$ denoising objective: $\mathcal{L}_{\text{diff}} = \left\| \epsilon_n - \hat{\epsilon}_n \right\|_2^2$.

\paragraph{Inference.} At test time, the model generates an entire action trajectory \( a_{1:T} \in \mathbb{R}^{T \times d} \) by iteratively denoising a noisy initialization. The process begins with Gaussian noise \( \tilde{a}^{(N)} \sim \mathcal{N}(0, I) \), and proceeds for \( N \) steps according to:
\[
\tilde{a}^{(n-1)} = \alpha_n \tilde{a}^{(n)} - \gamma_n \epsilon_\theta(\tilde{a}^{(n)}, z_{\text{pc}}, z_{\text{kp}}, c, n) + \sigma_n z_n,
\]
where \( z_n \sim \mathcal{N}(0, I) \) and \( \alpha_n, \gamma_n, \sigma_n \) are scheduler-defined scalars. The final output \( \tilde{a}^{(0)} \) represents the predicted trajectory over the full horizon.

\begin{figure*}[t]
  \centering
  \includegraphics[page=8, width=\linewidth]{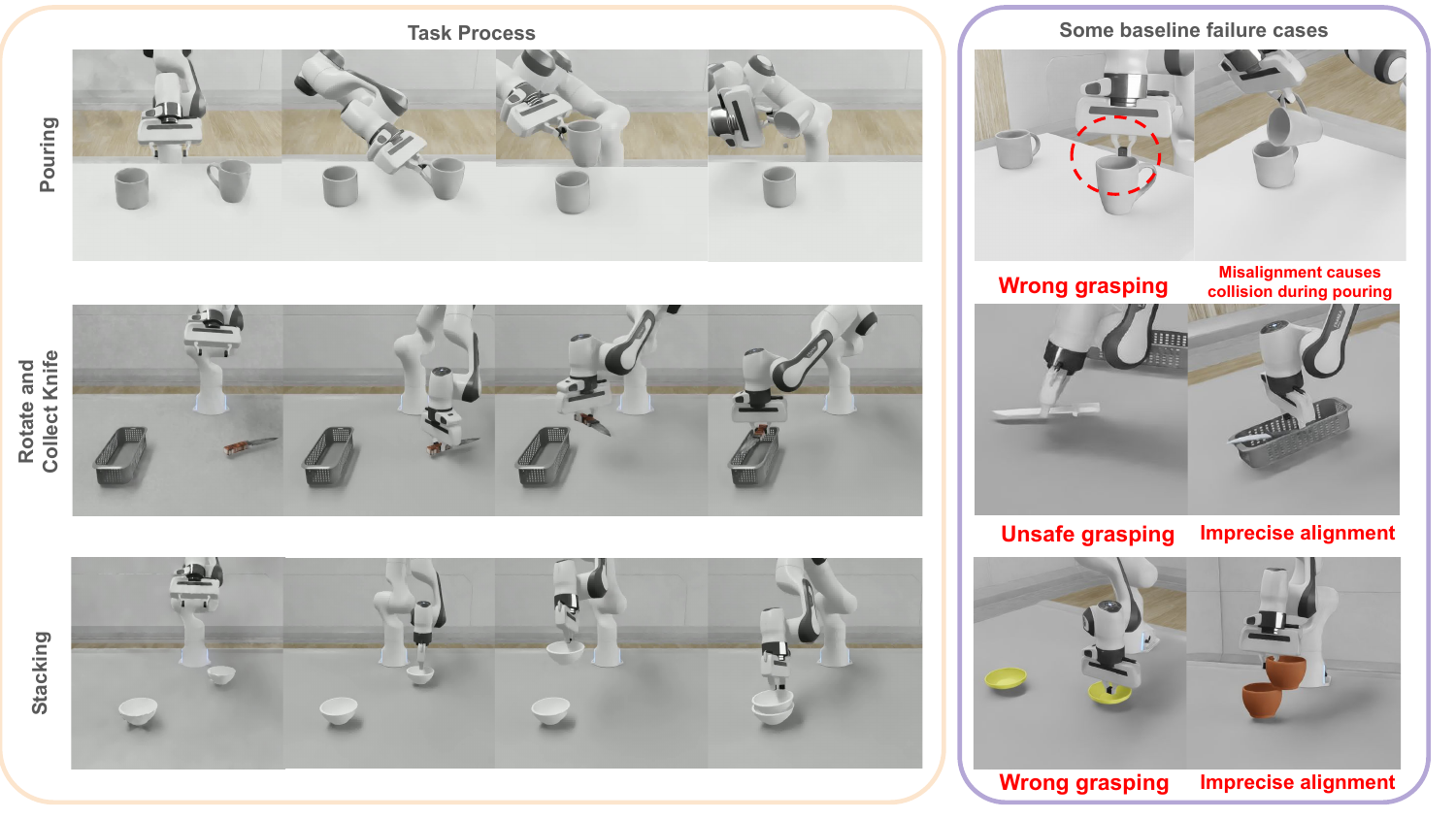}
  \caption{\textit{Left}: Representative rollout examples of our method on the three manipulation tasks. \textit{Right}: Common failure modes observed in baseline methods, including grasp misalignment, incorrect placement, and orientation errors.}
  \label{fig:qualitative result}
\end{figure*}

\section{Benchmark \& Demos}\label{sec:benchmark}
Existing benchmarks~\cite{james2020rlbench, liu2023libero, zhu2020robosuite, mees2022calvin} offer high-fidelity simulation but limited intra-category diversity and rarely evaluate \emph{pose-sensitive} control beyond pick-and-place. We therefore build a photorealistic Isaac Lab (Orbit)~\cite{mittal2023orbit} environment with three pose-sensitive tasks: \texttt{mug pouring}, \texttt{knife collecting}, and \texttt{bowl stacking}. Each episode instantiates ShapeNet objects~\cite{chang2015shapenetinformationrich3dmodel} by sampling instance, pose, and physical parameters, with optional support for scanned real-world meshes, and provides unified controls over camera intrinsics/extrinsics, object scale, backgrounds, textures, and lighting for domain randomization. Demonstrations are generated consistently across these variations by mapping a small set of canonical object-centric keyframe points to each instance and executing them with collision-checked IK and spline retiming.

\emph{Demo generation and recorded data.} Each task is generated by a compact FSM parameterized by object-centric geometric anchors computed from the CAD mesh in a canonical frame (e.g., handle/spout for mugs, blade axis for knives, rim axis for bowls). Anchors are mapped to world coordinates under the sampled pose, and the FSM produces SE(3) waypoints executed via collision-checked IK and spline retiming (with planner fallback), with guard conditions and success predicates; defining goals by anchors (not appearance) enables the same FSM to generalize across large shape and size variation. We record four-view RGB-D, fuse depth into object-centric point clouds, and log proprioception and actions, yielding paired observations and expert trajectories. We will open-source the environment, FSM pipeline, object packs/configs, and logging scripts with seeds and standardized evaluation protocols.

\section{Experiments}
We evaluate \textbf{\modelname} on three pose-sensitive manipulation tasks in simulation (\texttt{mug pouring}, \texttt{knife collecting}, \texttt{bowl stacking}) and study category-level generalization along three practical axes: \emph{instance transfer} to unseen objects under pose variation, \emph{data efficiency} as demonstrations per training object increase, and \emph{robustness to object scale shifts}. We compare against representative imitation and diffusion-policy baselines (ACT~\cite{act}, DP3~\cite{ze2024dp3}, 3DDA~\cite{3d_diffuser_actor}, BAKU~\cite{baku}, and P3-PO~\cite{levy2024p3poprescriptivepointpriors}) under the same sensing setup (multi-view RGB-D and proprioception) and training budget, and perform ablations to isolate the contributions of 3D keypoints and object-centric cropping. Beyond simulation, we additionally validate \modelname{} on real-world robot experiments for pushing-based shoe alignment and pouring, comparing to ACT, DP3, and GenDP; \modelname{} achieves consistently stronger performance, improving success by roughly 15\% over these baselines, demonstrating the practicality of our approach for real-world deployment.

\subsection{Experimental Setup}

\paragraph{Simulation.} We evaluate on three pose-sensitive tasks—\texttt{mug pouring}, \texttt{collect knife}, and \texttt{bowl stacking} (Fig.~\ref{fig:qualitative result}). For each task, we train a self-supervised 3D keypoint detector on 30 ShapeNet~\cite{chang2015shapenetinformationrich3dmodel} instances from the same category and feed the predicted keypoints to the policy. Policies are then trained with 100 self-generated trajectories on a single training instance per task and evaluated on 12 novel instances from the same category. Test objects are randomly positioned and oriented within a fixed region; we run 20 rollouts per object and report the mean success.

\paragraph{Real-World.} We further evaluate \modelname{} on real-world manipulation, including a pushing-based shoe alignment task and a pouring task. 
For each task, we train on two object instances (shoes or cups) with 60 trajectories per training object and test on three unseen objects. Since accurate depth sensing is challenging in the real world, we capture RGB images with 2 ZED cameras (fixed and wrist) and estimate depth using Depth-Anything V2~\cite{depth_anything_v2}, then use SAM2~\cite{sam} to segment the target object and reconstruct an object-centric point cloud. 
To mitigate sim-to-real gaps in keypoint detection, we finetune the keypoint extractor using additional cup and shoe meshes generated from web images. At evaluation time, we run 10 rollouts per unseen object and report the mean success rate.

\paragraph{Baselines.} In simulation, we compare against five representative methods: \textit{ACT}~\cite{act}, a 2D transformer policy using multi-view RGB; \textit{DP3}~\cite{ze2024dp3}, a 3D diffusion policy operating on fused point clouds; \textit{3D Diffuser Actor (3DDA)}~\cite{3d_diffuser_actor}, a point-cloud diffusion policy with 3D attention; \textit{BAKU}~\cite{baku}, a multi-task transformer with FiLM conditioning and action chunking; and \textit{P3\mbox{-}PO}~\cite{levy2024p3poprescriptivepointpriors}, which uses propagated point priors as the policy state. 
In real-world experiments, we compare to \textit{ACT}, \textit{DP3}, and \textit{GenDP}~\cite{wang2024gendp}, which is designed for object-level generalization. 
For fairness, all methods use the same sensing and training budget within each setting: 2D baselines consume multi-view RGB, while 3D baselines receive object-centric point clouds reconstructed using Depth-Anything V2~\cite{depth_anything_v2}; performance is reported by task success rate under matched evaluation protocols.

\begin{table}[t]
\centering
\small
\resizebox{\linewidth}{!}{
\begin{tabular}{lccccc}
\toprule
\multirow{2}{*}{Method} & \multicolumn{2}{c}{\textbf{Pouring}} & \multicolumn{2}{c}{\textbf{Collect Knife}} & \textbf{Stacking} \\
\cmidrule(lr){2-3}\cmidrule(lr){4-5}
 & Easy & Hard & Easy & Hard & (no pose var.) \\
\midrule
\rowcolor[HTML]{F7F7F7}
\multicolumn{6}{l}{\textbf{Seen Instances (in-distribution)}} \\
ACT~\cite{act}      & 85\%   & 35\%   & \textbf{95\%} & 35\%   & 30\% \\
\rowcolor[HTML]{EFEFEF}
DP3~\cite{ze2024dp3}                & 80\%   & 30\%   & 90\%          & 20\%   & 20\% \\
3DDA~\cite{3d_diffuser_actor}             & \textbf{95\%} & 40\%   & \textbf{95\%} & 45\%   & 65\% \\
\rowcolor[HTML]{EFEFEF}
BAKU~\cite{baku}                               & 90\%   & 55\%   & 80\%          & 45\%   & 60\% \\
P3-PO~\cite{levy2024p3poprescriptivepointpriors}                            & 90\%   & 65\%   & 85\%          & 50\%   & 55\% \\
\rowcolor[HTML]{EFEFEF}
\textbf{KeyGen (ours)}           & 90\%   & \textbf{90\%} & \textbf{95\%} & \textbf{85\%} & \textbf{85\%} \\
\midrule
\rowcolor[HTML]{F7F7F7}
\multicolumn{6}{l}{\textbf{Unseen Instances (out-of-distribution)}} \\
ACT~\cite{act}      & 35.4\% & 13.3\% & 75.8\% & 22.9\% & 22.1\% \\
\rowcolor[HTML]{EFEFEF}
DP3~\cite{ze2024dp3}                & 15\%   & 14.6\% & 54.5\% & 10.4\% & 18.8\% \\
3DDA~\cite{3d_diffuser_actor}             & 50.8\% & 20.0\% & \textbf{82.5\%} & 17.9\% & 57.9\% \\
\rowcolor[HTML]{EFEFEF}
BAKU~\cite{baku}                             & 40.8\% & 17.9\% & 63.8\% & 17.9\% & 35.0\% \\
P3-PO~\cite{levy2024p3poprescriptivepointpriors}                            & 42.9\% & 24.1\% & 61.7\% & 35.4\% & 42.4\% \\
\rowcolor[HTML]{EFEFEF}
\textbf{KeyGen (ours)}           & \textbf{68.3\%} & \textbf{56.3\%} & 77.5\% & \textbf{71.7\%} & \textbf{73.8\%} \\
\bottomrule
\end{tabular}
}
\vspace{4pt}
\caption{\textbf{Category-level generalization across tasks.} 
Success rates (\%) on both \emph{seen instances} (top block) and \emph{unseen instances} (bottom block). 
Each policy is trained on a single object and evaluated with 20 rollouts per configuration. 
\emph{Easy}: random position; \emph{Hard}: random position + orientation. \emph{Stacking} involves negligible pose variation.}
\label{tab:seen_unseen_results}
\end{table}

\begin{table}[t]
\centering
\small
\setlength{\tabcolsep}{4pt}
\resizebox{\linewidth}{!}{
\begin{tabular}{lcccccc}
\toprule
\multirow{2}{*}{Variant} & \multicolumn{2}{c}{\textbf{Pouring}} & \multicolumn{2}{c}{\textbf{Collect Knife}} & \multicolumn{2}{c}{\textbf{Stacking}} \\
\cmidrule(lr){2-3}\cmidrule(lr){4-5}\cmidrule(lr){6-7}
 & Seen & Unseen & Seen & Unseen & Seen & Unseen \\
\midrule
KeyGen w/o Object-centric    & 55\%  & 29.6\%  & 45\%  & 32.5\%  & 60\%  & 44.2\% \\
\rowcolor[HTML]{EFEFEF}
KeyGen w/o Keypoints         & \textbf{90\%}  & 45\%   & \textbf{90\%}  & 59.2\%  & \textbf{95\%}  & 67.5\% \\
KeyGen with 2D Keypoints     & 65\%  & 10.4\%  & 30\%  & 22.5\%  & 90\%  & 60\%   \\
\rowcolor[HTML]{EFEFEF}
\textbf{KeyGen (Full)}       & \textbf{90\%}  & \textbf{56.3\%}  & 85\%  & \textbf{71.7\%}  & 85\%  & \textbf{73.8\%} \\
\bottomrule
\end{tabular}}
\caption{\textbf{Ablation study}: success rates (\%) on seen and unseen objects across three tasks. We evaluate removing object-centric cropping, omitting keypoints, and replacing 3D keypoints with 2D image-based keypoints.}
\label{tab:ablation}
\end{table}

\begin{figure*}[t]
    \centering
    \includegraphics[width=1.0\linewidth]{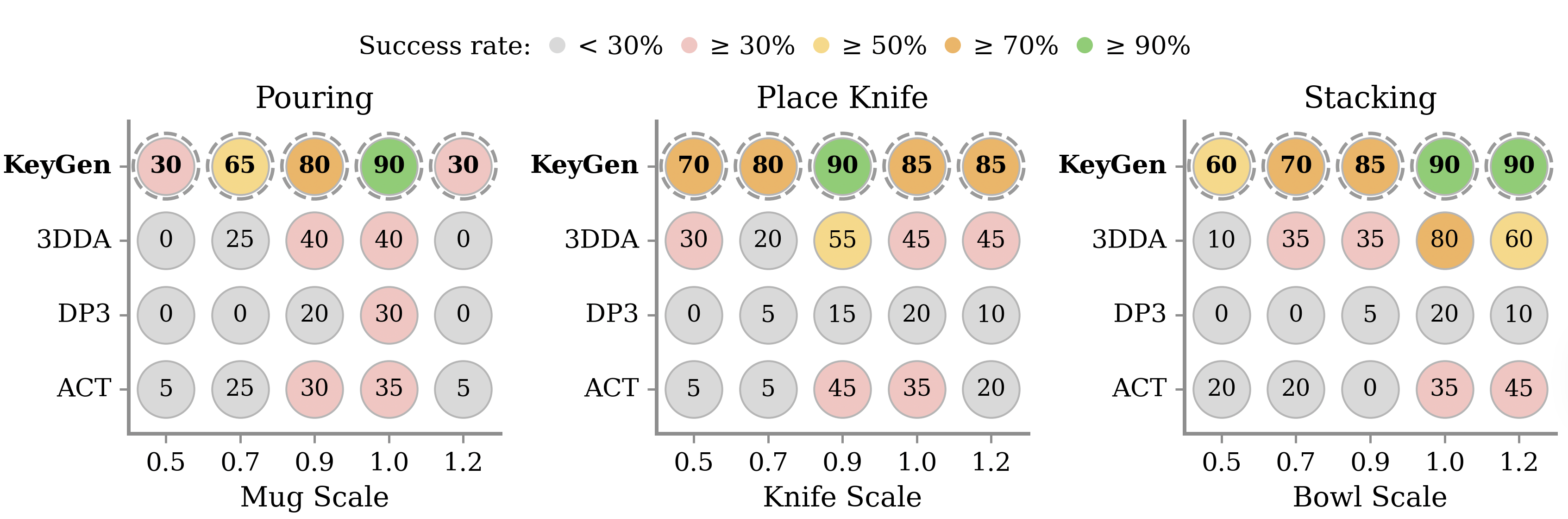}
    \caption{\textbf{Robustness to object-scale shifts} across three tasks. Policies are trained at scale 1.0 and evaluated zero-shot at isotropic multipliers \{0.5, 0.7, 0.9, 1.0, 1.2\} (x-axes). Each marker shows success rate (\%) for a method (rows); color encodes success-rate bins (legend). \modelname{} stays high under moderate rescaling and degrades mainly at the extremes, while baselines remain low or fluctuate.}
    \label{fig:scale}
\end{figure*}

\begin{figure}[!t]
  \centering
\includegraphics[width=1.0\linewidth]{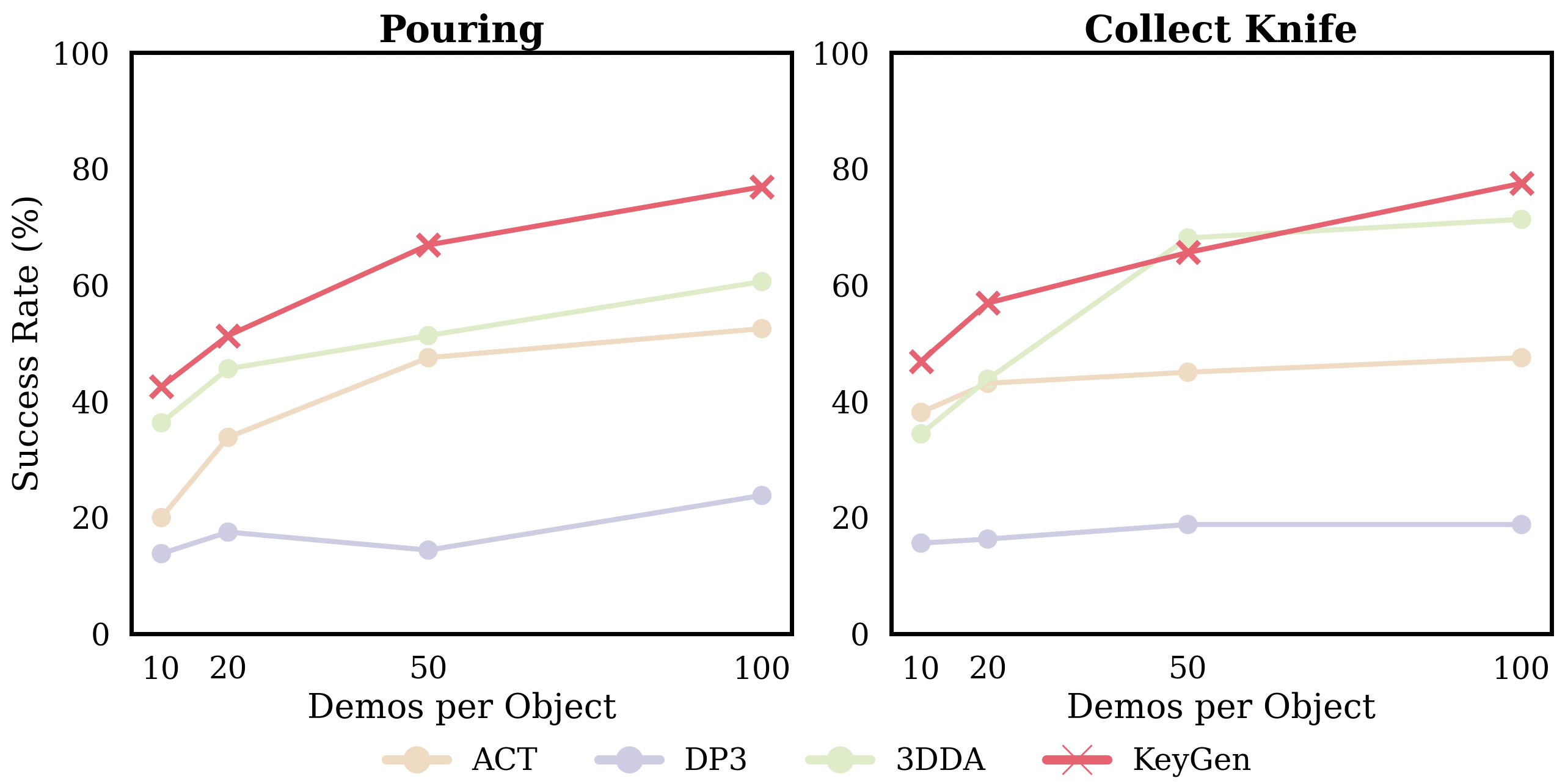}
  \caption{\textbf{Data efficiency} with more demonstrations per object. We fix the training object set and vary trajectories per object (10, 20, 50, 100). Curves show success on held-out objects for two tasks (\texttt{pouring}, \texttt{collect knife}). \modelname{} converts additional trajectories into steady gains and continues to improve at the highest data regime, while ACT grows slowly after early gains, 3DDA improves then flattens, and DP3 changes little.}
  \label{fig:data_efficiency}
\end{figure}

\subsection{Object-wise Generalization within a Category}
\label{subsec:object_generalization}
We evaluate whether a policy trained on few exemplars can execute the same task on novel objects from the same category with different shapes or poses. Each method is trained on a single object per task and evaluated on the seen training object and twelve unseen objects under an \emph{easy} setting (random translation) and a \emph{hard} setting (random translation + orientation), with results in Table~\ref{tab:seen_unseen_results}. On seen objects, all methods perform well in the easy setting (>\,85\% on pouring and knife), confirming task solvability, but in the hard setting baselines drop sharply while \modelname{} remains stable; this gap persists even for stacking, where pose variation is less critical, suggesting keypoint conditioning improves control fidelity beyond viewpoint robustness. The same pattern holds on unseen objects: under the hard setting, baselines frequently grasp the rim instead of the handle in pouring or fail to rotate into the slot for knife insertion, indicating reliance on appearance or scene layout, whereas \modelname{} anchors actions to consistent 3D keypoints (e.g., handle, spout, blade axis) that preserve functional geometry across instances, improving generalization on both seen and unseen objects.

\begin{figure*}[!t]
  \centering
  \includegraphics[width=1.0\linewidth]{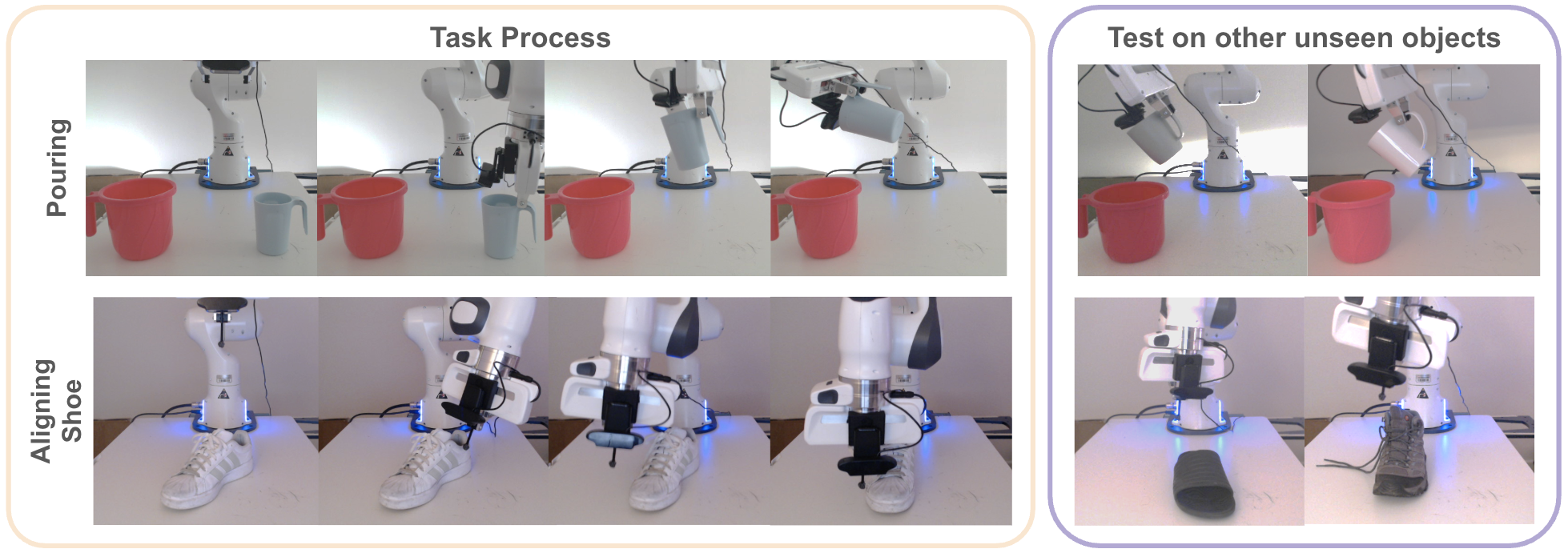}
  \caption{\textbf{Real-world qualitative results.} Left: execution snapshots for two real-robot tasks—\texttt{pouring} (top) and \texttt{align shoe} (bottom)—illustrating the task completion process. Right: rollouts on unseen object instances, demonstrating that \modelname{} generalizes across different cups and shoes with varying geometry and appearance.}
  \label{fig:realrobot}
\end{figure*}

\begin{figure}[!t]
  \centering
  \includegraphics[width=1.0\linewidth]{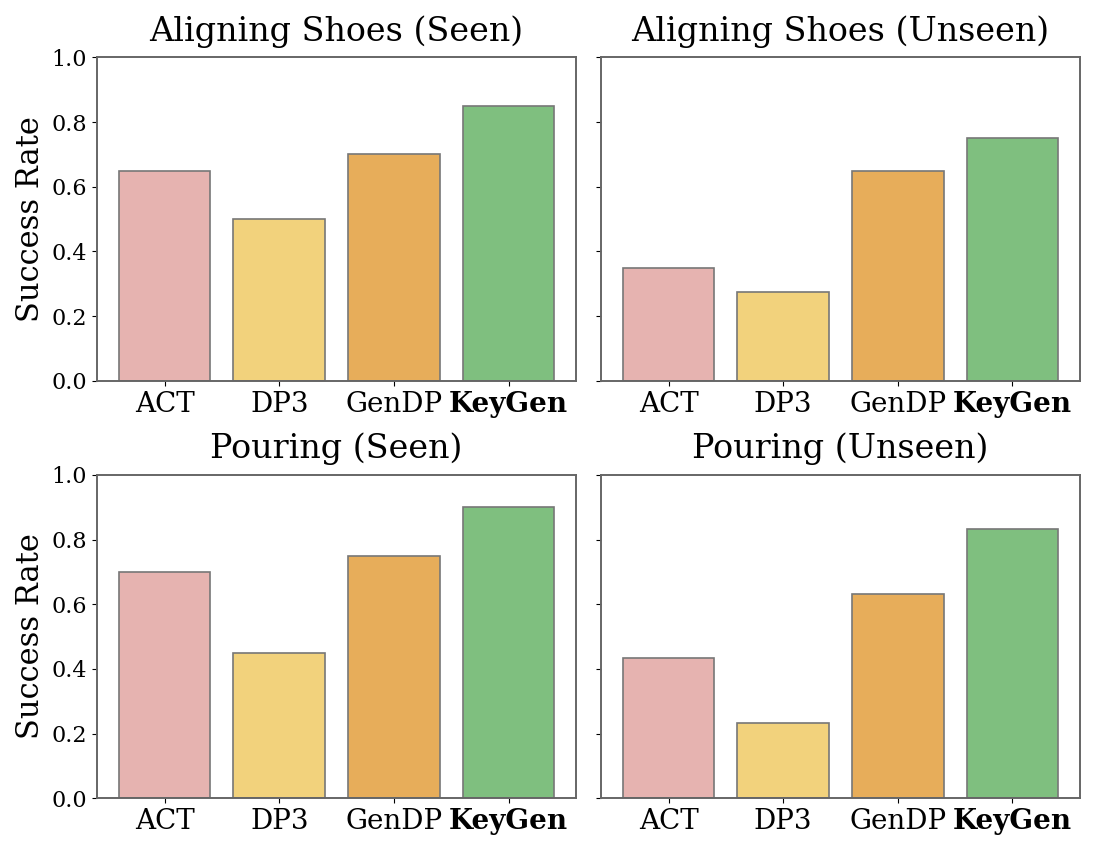}
  \caption{\textbf{Real-world performance on seen vs.\ unseen objects.} Success rates for two real-robot tasks—\texttt{align shoe} (top) and \texttt{pouring} (bottom)—evaluated on training instances (Seen, left) and held-out object instances (Unseen, right). \modelname{} achieves the highest success in both tasks and settings, outperforming ACT, DP3, and GenDP, with the largest gains on unseen objects.}
  \label{fig:barchart}
\end{figure}

\subsection{Scalability with Demonstrations per Object}
\label{subsec:scalability}
We evaluate data efficiency by fixing training object set (5 instances) and increasing demonstrations per object from 10 to 100, asking whether data improves category-level transfer rather than inducing overfitting; Figure~\ref{fig:data_efficiency} reports results for \texttt{pouring} and \texttt{collect knife} (with \texttt{stacking} following same trend). \modelname{} improves steadily as demonstrations increase and continues to gain even in highest-data regime, whereas ACT shows early gains then slows, 3DDA improves moderately before plateauing, and DP3 changes little. The widening gap suggests that canonicalized 3D keypoints provide correspondences across instances—so extra trajectories refine transferable geometric prior—while appearance-driven or unstructured point-cloud policies more plateau by fitting scene-specific cues.

\subsection{Robustness to Object-Scale Shifts}
\label{subsec:scale_robustness}
Because real objects vary in size, we train at nominal scale and evaluate zero-shot under isotropic rescaling across all three tasks (Figure~\ref{fig:scale}). \modelname{} remains robust to moderate scale shifts: pouring performs best near nominal scale and degrades mainly at extreme downsizing/upsizing where sensing and hardware limits dominate; knife insertion stays high across scales, indicating the policy preserves the required rotation and approach as blade thickness changes; and stacking remains strong at nominal and larger scales (often improving with larger bowls), while baselines are weaker or fluctuate. Overall, these results suggest that canonicalized 3D keypoints provide \emph{relative} geometric anchors (e.g., handle/spout, blade axis, rim) that transfer across size changes, enabling proportional adaptation of grasp and motion rather than reliance on absolute dimensions or appearance cues.

\subsection{Ablation Study}
To isolate the role of 3D keypoints and object-centric preprocessing, we vary the policy’s input representation (Table~\ref{tab:ablation}). Removing keypoints and training with only object-centric point clouds (w/o Keypoints) preserves strong performance on seen objects but drops noticeably on unseen ones, indicating that point clouds alone lack the structural consistency needed for transfer. Removing object-centric cropping (w/o Object-centric) further degrades performance, highlighting the importance of focusing on the relevant object region. Finally, replacing 3D keypoints with self-supervised 2D keypoints (Stable Keypoints~\cite{hedlin2024}) performs worse and incurs higher runtime, suggesting that 3D keypoints provide a more effective and efficient representation for category-level generalization than 2D alternatives.

\subsection{Real World Experiment Analysis}
Figure~\ref{fig:barchart} shows that \modelname{} achieves the highest success rates on both \texttt{pouring} and \texttt{align shoe}, consistently outperforming \textbf{ACT}~\cite{act}, \textbf{DP3}~\cite{ze2024dp3}, and \textbf{GenDP}~\cite{wang2024gendp} on both seen and unseen objects. \modelname{} reaches roughly \(0.85\!-\!0.90\) success on training instances and maintains \(0.75\!-\!0.85\) on held-out instances, while the best baseline typically drops to \(\sim 0.63\!-\!0.65\) on unseen objects, yielding an absolute gain of about \(15\%\) under instance transfer. We attribute this improvement to stronger object-centric geometric reasoning from semantic 3D keypoints: in pouring, \modelname{} reliably grasps the cup handle across geometries, whereas baselines often grasp visually salient but functionally incorrect regions (rim/side wall), causing unstable grasps and failed pours; in shoe alignment, keypoints provide an instance-agnostic pose cue that enables more consistent orientation alignment under shape/appearance variation, while baselines are more sensitive to instance-specific cues and misalign more often under distribution shift. Overall, keypoint-conditioned policies improve functional correspondence and pose awareness, translating into stronger real-world generalization.

\section{Conclusion}
\label{sec:conclusion}
\modelname{} is a diffusion-policy framework for category-level manipulation that generalizes to novel object instances by leveraging self-supervised 3D keypoints. By estimating a canonical orientation and predicting keypoints in a normalized frame, it produces semantically meaningful, pose-invariant object representations that serve as structured priors for action generation. Across three manipulation tasks, \modelname{} achieves strong performance on both seen and unseen objects, outperforming prior baselines and underscoring the value of transferable, task-agnostic keypoint representations for robotic control.

\section{Acknowledgment}
\label{sec:acknowledgment}
We are grateful to Ruoshi Liu for his insightful discussions, and to the PAIR lab members at Georgia Tech for their helpful feedback. This work was supported in part by the Stephen Fleming Early Career Grant, seed grants from IMS, Mechanical Engineering WIN, and IRIM at Georgia Tech, as well as GTRI.

\bibliographystyle{IEEEtran}
\bibliography{keygen,adapter}



\end{document}